\documentclass[preprint,12pt]{elsarticle}

\usepackage{hyperref}

\usepackage[utf8]{inputenc}
\usepackage{graphicx,amsfonts,amssymb,amsthm,stmaryrd,times}
\usepackage{amsmath}
\usepackage{graphicx,psfrag}
\usepackage[T1]{fontenc}
\usepackage[all]{xy}
\usepackage{multirow}
\usepackage{rotating}
\usepackage{xcolor}
\usepackage{dsfont}
\usepackage{wasysym}
\usepackage{xurl}
\usepackage{type1cm}
\usepackage{mathrsfs}
\usepackage{amsmath}
\usepackage{upgreek}
\usepackage{lettrine}
\usepackage[ruled,vlined]{algorithm2e}
\usepackage{booktabs}
\theoremstyle{definition}
\newtheorem{theorem}{Theorem}
\newtheorem*{theorem*}{Theorem}

\def\amax{\kern 0em\hbox{\rm \kern .25em\lower.1ex\hbox{\rlap{$\vee$}}\kern -.07em\lower.2ex\hbox{$\square$}\kern.25em}}
\def\amin{\kern 0em\hbox{\rm \kern .25em\lower.1ex\hbox{\rlap{$\wedge$}}\kern -.07em\lower.2ex\hbox{$\square$}\kern.25em}}

\def\boxmax{\kern 0em\hbox{\rm \kern .25em\lower.1ex\hbox{\rlap{$\vee$}}\kern -.07em\lower.2ex\hbox{$\square$}\kern.25em}}
\def\boxmin{\kern 0em\hbox{\rm \kern .25em\lower.1ex\hbox{\rlap{$\wedge$}}\kern -.07em\lower.2ex\hbox{$\square$}\kern.25em}}
\def\dualimp{\kern 0em\hbox{\rm \kern .25em\lower.1ex\hbox{\rlap{$\Rightarrow$}}\kern 0em\lower-1.2ex\hbox{$\overline{\hspace{2ex}}$}\kern.25em}}

\def\circmax{\kern 0em\hbox{\rm \kern .25em\lower.1ex\hbox{\rlap{$\vee$}}\kern -.18em\lower-.1ex\hbox{$\bigcirc$}\kern.25em}}
\def\circmin{\kern 0em\hbox{\rm \kern .25em\lower.1ex\hbox{\rlap{$\wedge$}}\kern -.18em\lower-.0ex\hbox{$\bigcirc$}\kern.25em}}

\newcommand{\sgn}{\text{sgn}}

\newcommand{\bb}{\begin{equation}}
\newcommand{\ee}{\end{equation}}

\newcommand{\R}{\mathbb{R}}
\newcommand{\F}{\mathbb{F}}
\newcommand{\Z}{\mathbb{Z}}
\newcommand{\E}{\mathbb{E}}
\newcommand{\G}{\mathbb{G}}
\newcommand{\B}{\mathbb{B}}
\newcommand{\N}{\mathbb{N}}

\newcommand{\M}{\mathbb{M}}
\newcommand{\A}{\boldsymbol{A}}
\newcommand{\X}{\boldsymbol{X}}
\newcommand{\Y}{\boldsymbol{Y}}

\newcommand{\vetu}{\boldsymbol{u}}
\newcommand{\vetx}{\boldsymbol{x}}
\newcommand{\vety}{\boldsymbol{y}}

\newcommand{\bpm}{\begin{bmatrix}}
\newcommand{\epm}{\end{bmatrix}}

\newcommand{\vetz}{\boldsymbol{z}}

\newcommand{\Q}{\mathcal{Q}}

\newcommand{\vett}{\boldsymbol{t}}

\newcommand{\argmax}[1]{\underset{#1}{\operatorname{argmax}}\,}

\begin{document}

\begin{frontmatter}


\title{Ensemble of Binary Classifiers Combined Using Recurrent Correlation Associative Memories}



\author{Rodolfo Anibal Lobo and Marcos Eduardo Valle}

\address{\textit{Institute of Mathematics, Statistics, and Scientific Computing} \\
\textit{University of Campinas}\\
Campinas, Brazil \\
valle@ime.unicamp.br, rodolfolobo@ug.uchile.cl
}

\begin{abstract}
An ensemble method should cleverly combine a group of base classifiers to yield an improved classifier. The majority vote is an example of a methodology used to combine classifiers in an ensemble method. In this paper, we propose to combine classifiers using an associative memory model. Precisely, we introduce ensemble methods based on recurrent correlation associative memories (RCAMs) for binary classification problems. We show that an RCAM-based ensemble classifier can be viewed as a majority vote classifier whose weights depend on the similarity between the base classifiers and the resulting ensemble method. More precisely, the RCAM-based ensemble combines the classifiers using a recurrent consult and vote scheme. Furthermore, computational experiments confirm the potential application of the RCAM-based ensemble method for binary classification problems.
\end{abstract}

\begin{keyword}
Binary classification \sep ensemble method  \sep associative memory \sep recurrent neural network \sep random forest.

\end{keyword}

\end{frontmatter}


\section{Introduction}

Inspired by the idea that multiple opinions are crucial before making a final decision, ensemble methods make predictions by consulting multiple different predictors  \cite{ponti2011combining}. Apart from their similarity with some natural decision-making methodologies, ensemble methods have a strong statistical background. Namely, ensemble methods aim to reduce the variance -- thus increasing the accuracy -- by combining multiple different predictors. Due to their versatility and effectiveness, ensemble methods have been successfully applied to a wide range of problems including classification, regression, and feature selection. As a preliminary study, this paper only addresses ensemble methods for binary classification problems.

Although there is no rigorous definition of an ensemble
 classifier \cite{kuncheva14}, they can be conceived as a group of base classifiers, also called weak or base classifiers. As to the construction of an ensemble classifier, we must take into account the diversity of the base classifiers and the rule used to combine them \cite{kuncheva14,polikar12chapter}. There are a plethora of ensemble methods in the literature, including bagging, pasting, random subspace, boosting, and stacking \cite{breiman96,Ho98,zhang12,geron2019hands}. For example, a bagging ensemble classifier is obtained by training copies of a single base classifier using different subsets of the training set \cite{breiman96}. Similarly, a random subspace classifier is obtained by training copies of a classifier using different subsets of features \cite{Ho98}. In both bagging and random subspace ensembles, the base classifiers are then combined using a voting scheme. Random forest is a successful example of an ensemble of decision tree classifiers trained using both bagging and random subspace ensemble ideas \cite{breiman01}.

In contrast to the traditional majority voting, in this paper, we propose to combine the base classifiers using an associative memory. Associative memories (AMs) refer to a broad class of mathematical models inspired by the human brain's ability to store and recall information by association \cite{austin87,kohonen87,hassoun97}. The Hopfield neural network is a typical example of a recurrent neural network able to implement an associative memory \cite{hopfield82}. Despite its many successful applications \cite{hopfield85,smith98,sun00,serpen08}, the Hopfield neural network suffers from an extremely low storage capacity as an associative memory model \cite{mceliece87}. To overcome the low storage capacity of the Hopfield network, many prominent researchers proposed alternative learning schemes \cite{kanter87,muezzinoglu05} as well as improved network architectures. In particular, the recurrent correlation associative memories (RCAMs), proposed by Chiueh and Goodman \cite{chiueh91}, can be viewed as a kernelized version of the Hopfield neural network  \cite{garcia04a,garcia04b,perfetti08}. In this paper, we apply the RCAMs to combine binary classifiers in an ensemble method.

At this point, we would like to remark that associative memories have been previously used by Kultur et al. to improve the performance of an ensemble method \cite{kultur2009ensemble}. Apart from addressing a regression problem, Kultur et al. use an associative memory in parallel to an ensemble of multi-layer perceptrons. The resulting model is called \textit{ensemble of neural networks with associative memory (ENNA)}. Our approach, in contrast, uses an associative memory to combine the base classifiers. Besides,  Kultur et al. associate patterns using the k-nearest neighbor algorithm which is formally a non-parametric method used for classification or regression. Differently, we use recurrent correlation associative memories, which are models conceived to implement associative memories.

The paper is organized as follows: The next section reviews the recurrent correlation associative memories. Ensemble methods are presented in Section \ref{sec:ensemble}. The main contribution of the manuscript, namely the ensemble classifiers based on associative memories, are addressed in Section \ref{sec:ensemble-AM}. Section \ref{sec:experiments} provides some computational experiments. The paper finishes with some concluding remarks in Section \ref{sec:concluding}.  

\section{A Brief Review on Recurrent Correlation Associative Memories} \label{sec:RCAMs}

Recurrent correlation associative memories (RCAMs) has been introduced by Chiueh and Goodman as an improved version of the famous correlation-based Hopfield neural network \cite{chiueh91,hopfield82}. 
 
Briefly, an RCAM is obtained by decomposing the Hopfield network with Hebbian learning into a two-layer recurrent neural network. The first layer computes the inner product (correlation) between the input and the memorized items followed by the evaluation of a non-decreasing continuous activation function. The subsequent layer yields a weighted average of the stored items.

In mathematical terms, a RCAM is defined as follows: Let $\mathbb{B} = \{-1,+1\}$ and $f:[-1,+1]\to \R$ be a continuous non-decreasing real-valued function. Given a fundamental memory set $\mathcal{U}=\{\vetu^1,\ldots,\vetu^P \} \subset \mathbb{B}^N$, the neurons in the first layer of a bipolar RCAM yield
\bb \label{eq:weights} w_\xi(t)= f\left(\frac{1}{N} \sum_{i=1}^N z_i(t) u_i^\xi\right), \quad \forall \xi \in 1,\ldots,P, \ee
where $\vetz(t)=[z_1(t),z_2(t),\ldots,z_N(t)]^T \in \mathbb{B}^N$ denotes the current state of the network and $\vetu^\xi = [u_1^\xi,\ldots,u_N^\xi]^T$ is the $\xi$th fundamental memory. The activation potential of the output neuron $a_i(t)$ is given by the following weighted sum of the memory items: 
\bb \label{eq:RCAM-activation} a_i(t)=\sum_{\xi=1}^P w_\xi(t) u_i^\xi, \quad \forall i=1,\ldots,N.\ee
Finally, the state of the $i$th neuron of the RCAM is updated as follows for all $i=1,\ldots,N$:
\bb \label{eq:RCAM-update}
z_i(t+1) = \begin{cases} \sgn\big(a_i(t)\big) & a_i(t) \neq 0, \\ z_i(t), & \mbox{otherwise}. \end{cases} \ee 
From \eqref{eq:RCAM-activation}, we refer to $w_\xi(t)$ as the weight associated to the $\xi$th memory item. 

In contrast to the Hopfield neural network, the sequence $\{\vetz(t)\}_{t \geq 0}$ produced by an RCAM is convergent in both synchronous and asynchronous update modes independently of the number of fundamental memories and the initial state vector $\vetz(0)$ \cite{chiueh91}. In other words, the limit $\vety = \lim_{t \to \infty} \vetz(t+1)$ of the sequence given by \eqref{eq:RCAM-update} is well defined using either synchronous or asynchronous update. 

As an associative memory model, an RCAM designed for the storage and recall of the vectors $\vetu^1,\ldots,\vetu^P$ proceeds as follows: Given a stimulus (initial state) $\vetz(0)$, the vector recalled by the RCAM is $\vety = \lim_{t \to \infty} \vetz(t+1)$.

Finally, the function $f$ defines different RCAM models. For example:
\begin{enumerate}
 \item The {\em correlation RCAM} or {\em identity RCAM} is obtained by considering in \eqref{eq:weights} the identity function $f_i(x)=x$.
 \item The {\em exponential RCAM}, which is determined by \bb \label{eq:f_e} f_e(x;\alpha)= e^{\alpha x}, \quad \alpha>0. \ee
\end{enumerate}
The identity RCAM corresponds to the traditional Hopfield network with Hebbian learning and self-feedback. Different from the Hopfield network and the identity RCAM, the storage capacity of the exponential RCAM scales exponentially with the dimension of the memory space. Apart from the high storage capacity, the exponential RCAM can be easily implemented on very large scale integration (VLSI) devices \cite{chiueh91}. Furthermore, the exponential RCAM allows for a Bayesian interpretation \cite{hancock98} and it is closely related to support vector machines and the kernel trick \cite{garcia04a,garcia04b,perfetti08}. In this paper, we focus on the exponential RCAM, formerly known as \textit{exponential correlation associative memory} (ECAM). 

\section{Ensemble of Binary Classifiers} \label{sec:ensemble}

An ensemble classifier combines a group of single classifiers, also called \textit{weak or base classifiers}, in order to provide better classification accuracy than a single one  \cite{ponti2011combining,zhang12,kuncheva14}. Although this approach is partially inspired by the idea that multiple opinions are crucial before making a final decision, ensemble classifiers have a strong statistical background. Namely, ensemble classifiers reduce the variance combining the base classifiers. Furthermore, when the amount of training data available is too small compared to the size of the hypothesis space, the ensemble classifier `` mixes'' the base classifiers reducing the risk of choosing the wrong single classifier \cite{kittler2003multiple}. 

Formally, let $ \mathcal{T} = \{(\vett_1,d_1),\dots, (\vett_M,d_M)\}$ be a training set where $\vett_i \in \mathcal{X}$ and $d_i \in \mathcal{C}$ are respectively the feature sample and the class label of the $i$th training pair. Here, $\mathcal{X}$ denotes the feature space and $\mathcal{C}$ represents the set of all class labels. In a binary classification problem, we can identify $\mathcal{C}$ with $\mathbb{B} = \{-1,+1\}$. Moreover, let $h_1,h_2,\ldots,h_P:\mathcal{X} \to \mathcal{C}$ be base classifiers trained using the whole or part of the training set $\mathcal{T}$.

Usually, the base classifiers are chosen according to their accuracy and diversity. On the one hand, an accurate classifier is one that has an error rate better than random guessing on new instances. On the other hand, two classifiers are diverse if they make different errors on new instances \cite{hansen1990neural,kittler2003multiple}.

Bagging and random subspace ensembles are examples of techniques that can be used to ensure the diversity of the base classifiers. The idea of bagging, an acronym for \textit{Bootstrap AGGregatING}, is to train copies of a certain classifier $h$ on subsets of the training set $\mathcal{T}$  \cite{breiman96}. The subsets are obtained by sampling the training $\mathcal{T}$ with replacement, a methodology known as \textit{bootstrap sampling} \cite{kuncheva14}. In a similar fashion, random subspace ensembles are obtained by training copies of a certain classifier $h$ using different subsets of the feature space \cite{Ho98}. Random forest, which is defined as an ensemble of decision tree classifiers, is an example of an ensemble classifier that combines both bagging and random subspace techniques \cite{breiman01}. 

Another important issue that must be addressed in the design of an ensemble classifier is how to combine the base classifiers. In the following, we review the majority voting methodology -- one of the oldest and widely used combination scheme. The methodology based on associative memories is introduced and discussed subsequently.

\subsection{Majority Voting Classifier}

As remarked by Kuncheva \cite{kuncheva14}, majority voting is one of the oldest strategies for decision making. In a wide sense, a majority voting classifier yields the class label with the highest number of occurrences among the base classifiers \cite{van2002overview,geron2019hands}.

Formally, let $h_1,h_2,\ldots,h_P:\mathcal{X} \to \mathcal{C}$ be the base classifiers. The \textit{majority {voting} classifier}, also called \textit{hard voting classifier} and denoted by $H_v:\mathcal{X} \to \mathcal{C}$, is defined by means of the equation
\begin{equation}
    \label{eq:hard_voting}
    H_v(\vetx) = \argmax{c \in \mathcal{C}} \sum_{\xi=1}^P w_\xi \mathcal{I}[h_\xi(\vetx)=c], \quad \forall \vetx \in \mathcal{X}, 
\end{equation}
where $w_1,\ldots,w_P$ are the weights of the base classifiers and $\mathcal{I}$ is the indicator function, that is,
\begin{equation}
\label{eq:indicator_function}
\mathcal{I}[h_\xi(\vetx)=c] = \begin{cases}
1, & h_\xi(\vetx)=c,\\ 
0, & \mbox{otherwise}.
\end{cases}
\end{equation}
When $\mathcal{C}=\{-1,+1\}$, the majority {voting} ensemble classifier given by \eqref{eq:hard_voting} can be written alternatively as
\begin{equation}
\label{eq:H_h}
    H_h(\vetx) = \sgn\left(\sum_{\xi=1}^P w_\xi h_\xi(\vetx)\right), \quad \forall \vetx \in \mathcal{X},
\end{equation} 
whenever $\sum_{\xi=1}^P w_\xi h_\xi(\vetx) \neq 0$ \cite{ferreira12chapter}.

\subsection{Ensemble Based on Bipolar Associative Memories} \label{sec:ensemble-AM}

Let us now introduce the ensemble classifiers based on the RCAM models. In analogy to the majority {voting} ensemble classifier, the RCAM-based ensemble classifier is formulated using only the base classifiers $h_1,\ldots,h_P:\mathcal{X} \to \mathbb{B}$. Precisely, consider a training set $\mathcal{T} = \{(\vett_i,d_i):i=1,\ldots,M\} \subset \mathcal{X}\times \mathbb{B}$ and let $X = \{\vetx_{1},\ldots,\vetx_{L} \} \subset \mathcal{X}$ be a batch of input samples. We first define the fundamental memories as follows for all $\xi=1,\ldots,P$:
\begin{equation}
\label{eq:bipolar-fundamental-memories}
\vetu^\xi = [h_\xi(\vett_1),\ldots,{h_\xi(\vett_M)},h_\xi(\vetx_1),\ldots,h_\xi(\vetx_L)]^T
\in \mathbb{B}^{M+L}.
\end{equation}
In words, the $\xi$th fundamental memory is obtained by concatenating the outputs of the $\xi$th base classifier evaluated at the $M$ training samples and the $L$ input samples. The bipolar RCAM is synthesized using the fundamental memory set $\mathcal{U}=\{\vetu^1,\ldots,\vetu^P\}$ and it is initialized at the state vector
\begin{equation}
    \label{eq:stimulus}
     \vetz(0) = [d_1,d_2,\ldots,d_M, \underbrace{0,0,\ldots,0}_{L-\mbox{\small times}}]^T.
\end{equation}
Note that the first $M$ components of initial state $\vetz(0)$ correspond to the targets in the training set $\mathcal{T}$. The last $L$ components of $\vetz(0)$ are zero, a neutral element different from the class labels. The inital state $\vetz(0)$ is presented as input to the associative memory and the last $L$ components of the recalled vector $\vety$ yield the class label of the batch of input samples $X = \{\vetx_1,\ldots,\vetx_L\}$. In mathematical terms, the RCAM-based ensemble classifier $H_a:\mathcal{X} \to \mathbb{B}$ is defined by means of the equation
\begin{align}
\label{eq:H_a1}
    H_a(\vetx_i) = y_{M+i}, \quad \forall \vetx_i \in X,
\end{align}
where $\vety = [y_1,\ldots,y_M,y_{M+1},\ldots,y_{M+L}]^T$ is the limit of the sequence $\{\vetz(t)\}_{t\geq 0}$  given by \eqref{eq:RCAM-update}. 

In the following, we point out the relationship between the bipolar RCAM-based ensemble classifier and the majority voting ensemble described by \eqref{eq:H_h}. Let $\vety$ be the vector recalled by the RCAM fed by the input $\vetz(0)$ given by \eqref{eq:stimulus}, that is, $\vety$ is a stationary state of the RCAM.
From \eqref{eq:RCAM-activation}, \eqref{eq:RCAM-update}, and \eqref{eq:bipolar-fundamental-memories}, the output of the RCAM-based ensemble classifier satisfies 
\bb \label{eq:H_a}
H_a(\vetx_i) = \sgn \left(\sum_{\xi=1}^P w_\xi h_\xi(\vetx_i) \right), \ee
where 
\bb w_\xi = f\left(\frac{1}{M+L}\sum_{i=1}^{M+L} y_i u_i^\xi\right), \quad \forall \xi=1,\ldots,P. \ee 
From \eqref{eq:H_a}, the bipolar RCAM-based ensemble classifier can be viewed as a weighted majority {voting} classifier. Furthermore, the weight $w_\xi$ depends on the similarity between the $\xi$th base classifier $h_\xi$ and the ensemble classifier $H_a$. Precisely, let us define the similarity between two binary classifiers $H,h_\xi:\mathcal{X} \to \mathbb{B}$ on a set of samples $S$ by means of the equation 
\bb \label{eq:similarity_Hh}
\mathtt{Sim}(H,h) = \frac{1}{\mbox{Card}(S)} \sum_{\boldsymbol{s} \in S} \mathcal{I}\big[ h(\boldsymbol{s}) = H(\boldsymbol{s})\big].
\ee
Using \eqref{eq:similarity_Hh}, we can state the following theorem:
\begin{theorem} \label{main-theorem}
The weights of the RCAM-based ensemble classifier given by \eqref{eq:H_a} satisfies the following identities for all $\xi=1,\ldots,P$:

\bb \label{eq:thm_wt} w_\xi = f \big(1-2\cdot \mathtt{Sim}(H_a, h_\xi) \big), \quad \forall t \geq 1, \ee
where the similarity in \eqref{eq:thm_wt} is evaluated on the union of all training and input samples, that is, on $S = X \cup T = \{\vett_1,\ldots,\vett_M\}\cup\{\vetx_1,\ldots,\vetx_L\}$.
\end{theorem}

\begin{proof}
Since we are considering a binary classification problem, the similarity between the ensemble $H_a$ and the base classifier $h_\xi$ on $S = X \cup T$, with $N = \text{Card}(S) = M+L$, satisfies the following identities:
\begin{align*}
    \mathtt{Sim}(H, h) &= 
    1 - \frac{1}{N}\sum_{i=1}^N \mathcal{I}[h(\boldsymbol{s}_i) \neq H_a(\boldsymbol{s}_i)] 
    = 1 - \frac{1}{4N}\sum_{i=1}^N \big(h(\boldsymbol{s}_i)- H_a(\boldsymbol{s}_i)\big)^2 \\
    &= 1 - \frac{1}{2N}\sum_{i=1}^N \big(1-H_a(\boldsymbol{s}_i) h(\boldsymbol{s}_i)\big) 
    = \frac{1}{2}\left(1 - \frac{1}{N} \sum_{i=1}^N H_a(\boldsymbol{s}_i) h(\boldsymbol{s}_i) \right)
\end{align*}
Equivalently, we have 
{
\bb  \label{eq:Sim}
\frac{1}{\mbox{Card}(S)} \sum_{\boldsymbol{s} \in S} H(\boldsymbol{s})h(\boldsymbol{s}) = 1 - 2 \cdot \mathtt{Sim}(H,h).
\ee }
Now, from \eqref{eq:weights}, \eqref{eq:H_a1}, and \eqref{eq:Sim}, we obtain the following identities: 
{
\begin{align*} w_\xi &= f\left(\frac{1}{N}\sum_{i=1}^N y_i u_i^\xi \right) = f\big(1-2\cdot \mathtt{Sim}(H_a,h_\xi)\big),
\end{align*}}
which concludes the proof.
\end{proof}

Theorem \ref{main-theorem} shows that the RCAM-based ensemble classifier is a majority {voting} classifier whose weights depend on the similarity between the {base classifiers} and the ensemble itself. In fact, in view of the dynamic nature of the RCAM model, $H_a$ is obtained by a recurrent consult and vote scheme. Moreover, at the first step, the weights depend on the accuracy of the base classifiers.

\section{Computational Experiments} \label{sec:experiments}

In this section, we perform some computational experiments to evaluate the performance of the proposed RCAM-based ensemble classifiers for binary classification tasks. 
Precisely, we considered the RCAM-based ensembles obtained using the identity and the exponential as the activation function $f$.  The parameter $\alpha$ of the exponential activation function has been either set to $\alpha=1$ or it has been determined using a grid search on the set $\{10^{-2},10^{-1},0.5,1,5,10,20,50\}$ with 5-fold cross-validation on the training set. 
The RCAM-based ensemble classifiers have been compared with AdaBoost, gradient boosting, and random forest ensemble classifiers, all available at the python's \texttt{scikit-learn API} (\texttt{sklearn}) \cite{scikit-learn}. 

First of all, we trained AdaBoost and gradient boosting ensemble classifiers using the default parameters of \texttt{sklearn}. Recall that boosting ensemble classifiers are developed incrementally by adding base classifiers to reduce the number of misclassified samples \cite{kuncheva14}. Also, we trained the random forest classifier with 30 base classifiers ($P=30$) \cite{breiman01}. Recall that the base classifiers of the random forest are decision trees obtained using bagging and random subspace techniques \cite{breiman96,Ho98}. Then, we used the base classifiers from the trained random forest ensemble to define the RCAM-based ensemble. In other words, the same base classifiers $h_1,\ldots,h_{30}$ are used in the random forest and the RCAM-based classifiers. The difference between the ensemble classifiers resides in the combining rule. Recall that the random forest combines the base classifiers using majority voting. From the computational point of view, training the random forest and the RCAM-ensemble classifiers required similar resources. Moreover, despite the consult and vote scheme of the RCAM-based ensemble, they have not been significantly more expensive than the random forest classifier. The grid search used to fine-tune the parameter $\alpha$ of the exponential RCAM-based ensemble is the major computational burden in this computational experiment.      

For the comparison of the ensemble classifiers, we considered {28} binary classification problems from the OpenML repository \cite{OpenML}. These binary classification problems can be obtained using the command \texttt{fetch\_openml} from \texttt{sklearn}. We would like to point out that missing data has been handled before splitting the data set into training and test sets using the command \texttt{SimpleImputer} from \texttt{sklearn}. Also, we pre-processed the data using the \texttt{StandardScaler} transform. Therefore, each feature is normalized by subtracting the mean and dividing by the standard deviation, both computed using only the training set. Furthermore, since some data sets are unbalanced, we used the F-measure to evaluate quantitatively the performance of a certain classifier. 
Table \ref{tab:accuracies} shows the mean and the standard deviation of the F-measure obtained from the ensemble classifiers using stratified 10-fold cross-validation. The largest F-measures for each data set have been typed using boldface.

\begin{table*}[t]
	\caption{Mean and standard deviation of the F-measures produced by ensemble classifiers using stratified 10-fold cross-validation.} \label{tab:accuracies}
	\centering \resizebox{\textwidth}{!}{
	\begin{tabular}{|l||l|l|l|l|l|l|}
		\toprule
		&  & \textbf{Gradient} &
		 \textbf{Random} & \textbf{Identity} & \textbf{Exponential} & \textbf{Exp. RCAM} \\
		\textbf{Data set} & \textbf{AdaBoost} & \textbf{Boosting} & \textbf{Forest} & \textbf{RCNN} & \textbf{RCAM} $\qquad$ & \textbf{+ Grid Search} \\ \midrule
Arsene & $84.0\pm5.9$ & $\mathbf{86.2\pm7.6}$ & $81.5\pm8.9$ & $83.8\pm8.4$ & $83.8\pm8.4$ & $85.2\pm10.2$ \\ 
Australian & $82.1\pm3.4$ & $\mathbf{85.8\pm3.8}$ & $85.4\pm3.4$ & $85.3\pm2.9$ & $85.3\pm2.9$ & $85.0\pm2.9$ \\ 
Banana & $67.9\pm2.1$ & $88.1\pm1.6$ & $88.0\pm1.3$ & $\mathbf{88.2\pm1.2}$ & $88.2\pm1.2$ & $87.2\pm1.2$ \\ 
Banknote & $\mathbf{99.6\pm0.4}$ & $99.5\pm0.9$ & $99.3\pm0.7$ & $99.2\pm0.7$ & $99.2\pm0.7$ & $98.9\pm0.9$ \\ 
Blood Transfusion & $\mathbf{43.0\pm13.1}$ & $37.9\pm11.2$ & $32.3\pm10.4$ & $33.3\pm10.6$ & $33.3\pm10.6$ & $32.5\pm8.2$ \\ 
Breast Cancer Wisconsin & $94.7\pm2.0$ & $95.2\pm2.4$ & $94.9\pm3.4$ & $\mathbf{95.4\pm2.9}$ & $95.1\pm3.3$ & $95.2\pm4.2$ \\ 
Chess & $96.5\pm1.1$ & $97.9\pm0.8$ & $99.0\pm0.5$ & $99.0\pm0.6$ & $99.0\pm0.6$ & $\mathbf{99.2\pm0.4}$ \\ 
Colic & $87.1\pm6.4$ & $86.7\pm7.4$ & $88.7\pm5.7$ & $88.6\pm5.4$ & $88.6\pm5.4$ & $\mathbf{88.9\pm4.6}$ \\ 
Credit Approval & $86.4\pm2.9$ & $86.9\pm3.2$ & $88.4\pm2.8$ & $\mathbf{88.4\pm2.5}$ & $\mathbf{88.4\pm2.5}$ & $88.3\pm2.3$ \\ 
Credit-g & $82.3\pm2.5$ & $84.2\pm2.8$ & $83.7\pm2.4$ & $\mathbf{84.3\pm2.2}$ & $\mathbf{84.3\pm2.2}$ & $83.9\pm1.8$ \\ 
Cylinder Bands & $78.3\pm4.8$ & $84.0\pm4.8$ & $83.0\pm6.6$ & $83.3\pm6.4$ & $83.3\pm6.4$ & $\mathbf{87.0\pm4.2}$ \\ 
Diabetes & $63.1\pm5.2$ & $65.1\pm6.5$ & $63.9\pm8.8$ & $\mathbf{65.6\pm8.2}$ & $\mathbf{65.6\pm8.2}$ & $62.4\pm7.8$ \\ 
Egg-Eye-State & $70.1\pm1.3$ & $78.0\pm0.9$ & $91.5\pm0.7$ & $91.8\pm0.8$ & $91.8\pm0.8$ & $\mathbf{92.9\pm0.8}$ \\ 
Haberman & $\mathbf{35.4\pm9.5}$ & $30.8\pm14.2$ & $27.4\pm13.4$ & $30.6\pm9.6$ & $30.6\pm9.6$ & $34.9\pm12.9$ \\
Hill-Valley & $40.9\pm5.4$ & $52.9\pm7.3$ & $54.9\pm4.6$ & $56.6\pm3.8$ & $56.6\pm4.0$ & $\mathbf{59.1\pm6.2}$ \\ 
Internet Advertisements & $98.0\pm0.3$ & $98.6\pm0.3$ & $\mathbf{98.8\pm0.4}$ & $98.7\pm0.4$ & $98.7\pm0.4$ & $98.7\pm0.5$ \\ 
Ionosphere & $94.3\pm1.7$ & $94.4\pm2.0$ & $94.2\pm2.5$ & $94.0\pm2.5$ & $94.0\pm2.5$ & $\mathbf{94.7\pm2.7}$ \\ 
MOFN-3-7-10 & $\mathbf{100.0\pm0.0}$ & $\mathbf{100.0\pm0.0}$ & $99.8\pm0.2$ & $99.7\pm0.3$ & $99.7\pm0.3$ & $99.7\pm0.5$ \\
Monks-2 & $0.0\pm0.0$ & $69.3\pm8.7$ & $93.1\pm3.3$ & $93.5\pm3.3$ & $93.5\pm3.3$ & $\mathbf{98.5\pm2.7}$ \\ 
Phoneme & $68.3\pm3.0$ & $75.4\pm2.4$ & $84.0\pm3.0$ & $84.1\pm2.7$ & $84.1\pm2.7$ & $\mathbf{85.7\pm2.0}$ \\ 
Pishing Websites & $94.4\pm0.4$ & $95.3\pm0.5$ & $97.5\pm0.6$ & $97.4\pm0.6$ & $97.4\pm0.6$ & $\mathbf{97.5\pm0.5}$ \\ 
Sick & $78.3\pm6.4$ & $88.8\pm3.9$ & $87.5\pm3.1$ & $88.6\pm3.9$ & $88.6\pm3.9$ & $\mathbf{89.7\pm3.6}$ \\ 
Sonar & $\mathbf{83.9\pm8.0}$ & $81.3\pm6.2$ & $81.9\pm11.4$ & $83.3\pm11.1$ & $83.3\pm11.1$ & $83.2\pm11.1$ \\ 
Spambase & $91.8\pm1.5$ & $93.1\pm1.7$ & $\mathbf{94.2\pm1.1}$ & $94.0\pm1.2$ & $94.1\pm1.2$ & $94.0\pm1.2$ \\ 
Steel Plates Fault & $\mathbf{100.0\pm0.0}$ & $\mathbf{100.0\pm0.0}$ & $99.0\pm0.8$ & $99.2\pm0.6$ & $99.2\pm0.6$ & $99.4\pm0.7$ \\ 
Tic-Tac-Toe & $84.5\pm2.6$ & $94.8\pm2.1$ & $95.6\pm1.2$ & $95.5\pm1.2$ & $95.5\pm1.2$ & $\mathbf{96.5\pm1.5}$ \\ 
Titanic & $\mathbf{58.8\pm4.3}$ & $53.8\pm4.4$ & $53.6\pm4.2$ & $53.6\pm4.2$ & $53.6\pm4.2$ & $53.8\pm4.4$ \\ 
ilpd & $\mathbf{41.4\pm11.4}$ & $35.3\pm15.1$ & $35.1\pm15.8$ & $37.5\pm16.6$ & $37.5\pm16.6$ & $33.5\pm14.6$ \\  
		\bottomrule
	\end{tabular} }
\end{table*}
Note {the exponential} RCAM-based ensemble classifier with grid search produced the largest F-measures in 11 of the {28} data sets. In particular, the exponential RCAM with grid search produced outstanding F-measures on the ``Monks-2'' and  ``Egg-Eye-State'' data sets. For a better comparison of the ensemble classifiers, we followed Dem\v{s}ar's recommendations to compare multiple classifier models using multiple data sets \cite{Demsar06}. The Friedman test rejected the hypothesis that there is no difference between the ensemble classifiers. 

\begin{figure}[t] 
    \centering
    \includegraphics[width=1\columnwidth]{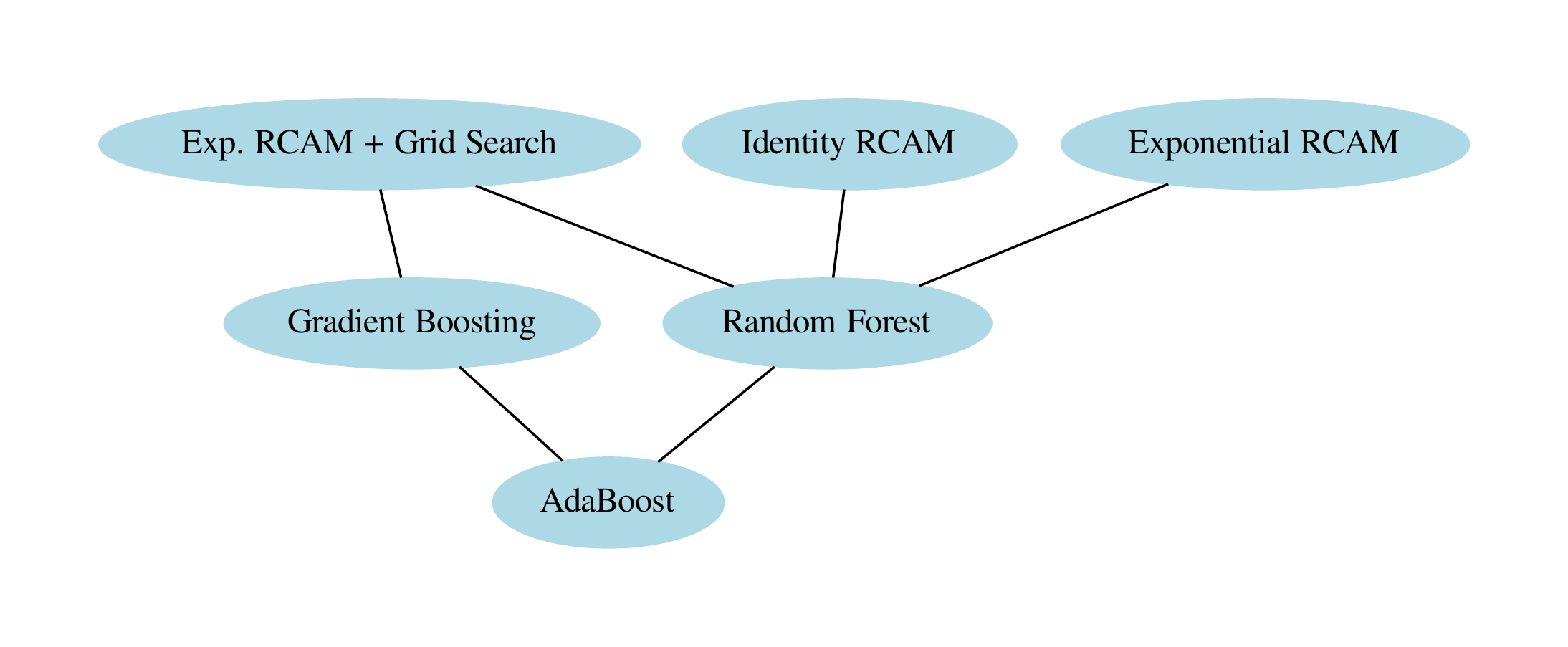}
    \caption{Hasse diagram of Wilcoxon signed-rank test with a confidence level at 95\%.}
    \label{fig:hotdiagram}
\end{figure}

A visual interpretation of the outcome of this computational experiment is provided in Figure \ref{fig:hotdiagram} with the Hasse diagram of the non-parametric Wilcoxon signed-rank test with a confidence level at 95\% \cite{burda13,weise15}. In this diagram, an edge means that the classifier on the top statistically outperformed the classifier on the bottom. The outcome of this analysis confirms that the RCAM-based ensemble classifiers statistically outperformed the other ensemble methods: AdaBoost, gradient boosting, and random forest. 

\begin{figure}[t] 
    \centering
    \includegraphics[width=1\columnwidth]{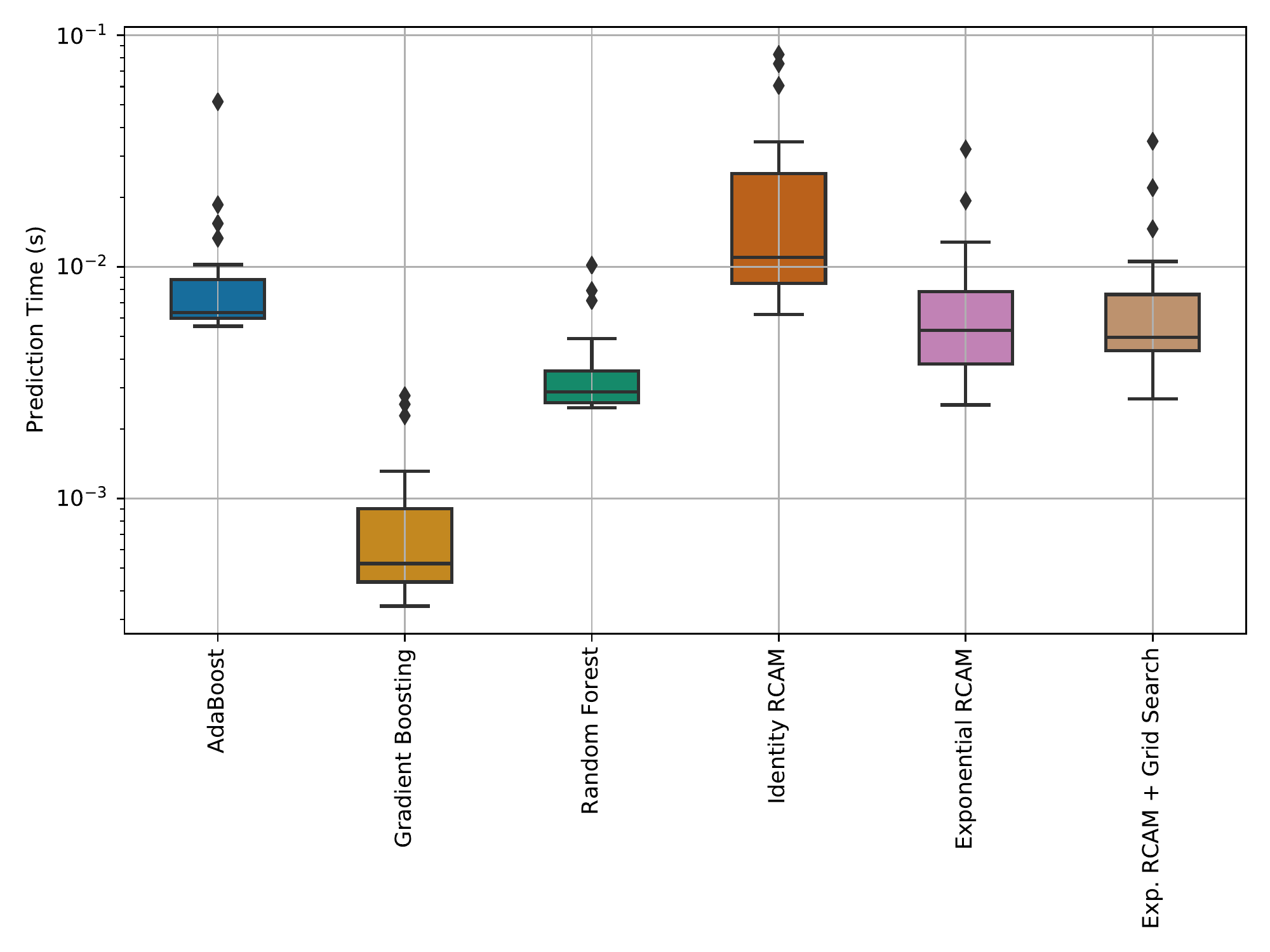}
    \caption{Box-plot of the average time for prediction of batch of input samples.}
    \label{fig:boxplot_time}
\end{figure}

As to the computational effort, Figure \ref{fig:boxplot_time} shows the average time required by the ensemble classifiers for the prediction of a batch of testing samples. Note that the most expensive method is identity RCAM-based ensemble classifier while the gradient boosting is the cheapest. The exponential RCAM-based ensemble is less expensive than the AdaBoost and quite comparable to the random forest classifier.

\begin{figure}[t] 
    \centering
    \includegraphics[width=1\columnwidth]{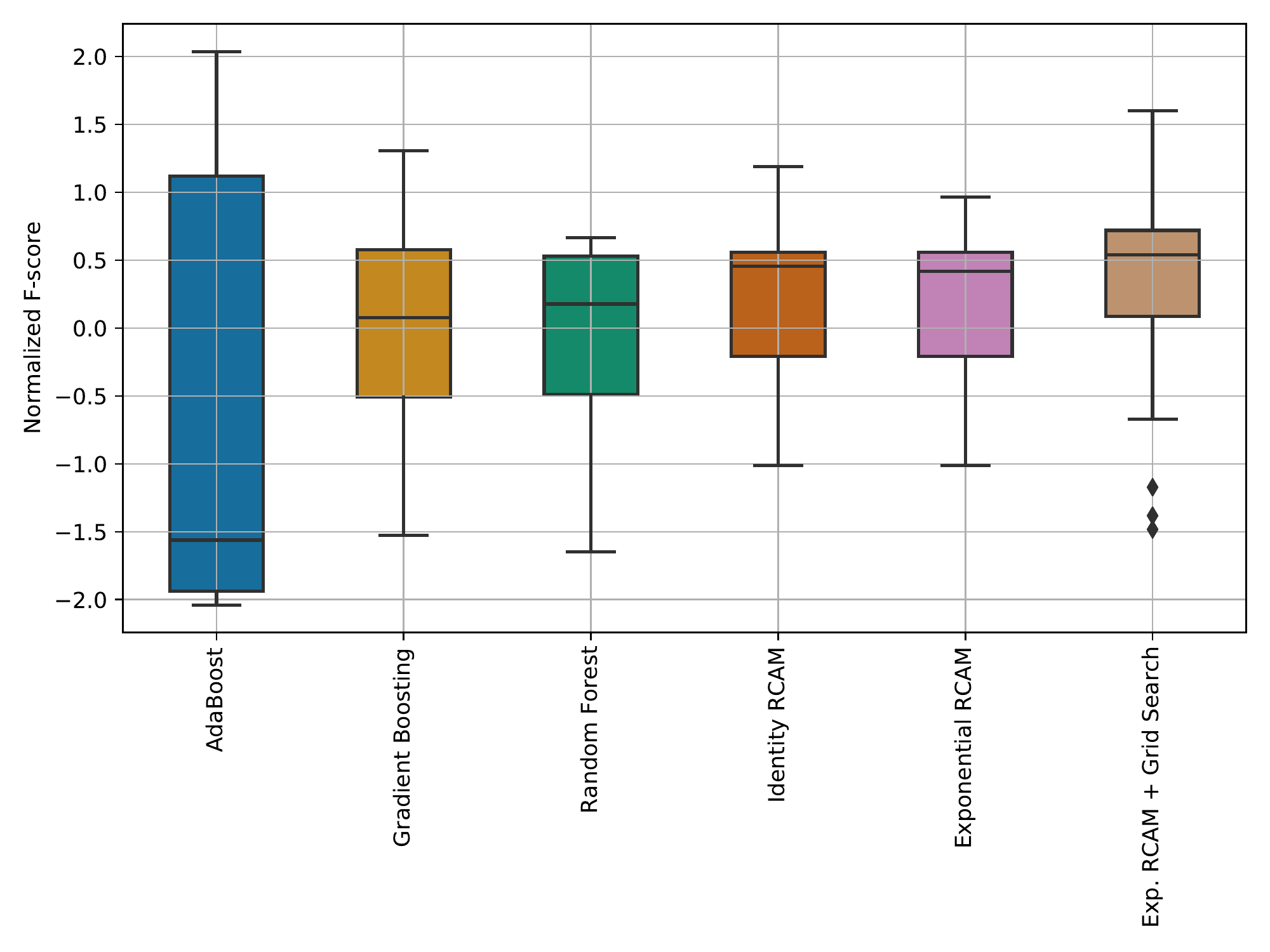}
    \caption{Box-plot of the normalized F-measures produced by the ensemble classifiers.}
    \label{fig:boxplot}
\end{figure}

Finally, note from Table \ref{tab:accuracies} that some problems such as the ``Banknote'' and the ``MOFN-3-7-10'' data sets are quite easy while others such as the  ``Haberman'' and ``Hill Valey'' are very hard. In order to circumvent the difficulties imposed by each data set, Figure \ref{fig:boxplot} shows a box-plot with the normalized F-measure values provided in Table \ref{tab:accuracies}. Precisely, for each data set (i.e., each row in Table \ref{tab:accuracies}), we subtracted the mean and divided by the standard deviation of the score values. 

The box-plot in Figure \ref{fig:boxplot} confirms the good performance of the RCAM-based ensemble classifiers, including the exponential RCAM-based ensemble classifier with a grid search. Concluding, the boxplots shown on Figures \ref{fig:boxplot_time} and \ref{fig:boxplot} supports the potential application of the RCAM models as an ensemble of classifiers for binary classification problems.

\section{Concluding Remarks}  \label{sec:concluding}

This paper provides a bridge between ensemble methods and associative memories. In general terms, an ensemble method reduces variance and improve the accuracy and robustness by combining a group of base predictors \cite{zhang12,kuncheva14}. The rule used to combine the base predictors {is one important} issue in the design of an ensemble method. In this paper, we propose to combine the base predictors using an associative memory. Associative memory is a model designed for the storage and recall of a set of vectors \cite{hassoun97}. Furthermore, an associative memory should be able to retrieve {a stored} item from a corrupted or partial version of it. In an ensemble method, the memory model is designed for the storage of evaluations of the base classifiers. The associative memory is then fed by a vector with the target of training data as well as the unknown predictions. The output of the ensemble method is obtained from the vector retrieved by the memory.

Specifically, in this paper, we presented ensemble methods based on the recurrent correlation associative memories (RCAMs) for binary classifications. RCAMs, proposed by Chiueh and Goodman \cite{chiueh91}, are high storage capacity associative memories which, besides Bayesian and kernel trick interpretation, are particularly suited for VLSI implementation \cite{hancock98,garcia04a,garcia04b,perfetti08}. Theorem \ref{main-theorem} shows that the RCAM model yields a majority {voting} classifier whose weights are obtained by a recurrent consult and vote scheme. Moreover, the weights depend on the similarity between the {base classifiers} and the resulting ensemble. Computational experiments using decision tree as the base classifiers revealed an outstanding performance of the exponential RCAM-based ensemble classifier combined with a grid search strategy to fine-tune its parameter. The exponential RCAM-based ensemble, in particular, outperformed the traditional AdaBoost, gradient boosting, and random forest classifiers. 

In the future, we plan to investigate further associative memory-based ensemble methods. In particular, we plan to extend these ensemble methods to multi-class classification problems using, for instance, multistate associative memory models \cite{jankowski96,muezzinoglu03,minemoto16,kobayashi17a}.


\begin{thebibliography}{38}
\expandafter\ifx\csname natexlab\endcsname\relax\def\natexlab#1{#1}\fi
\providecommand{\url}[1]{\texttt{#1}}
\providecommand{\href}[2]{#2}
\providecommand{\path}[1]{#1}
\providecommand{\DOIprefix}{doi:}
\providecommand{\ArXivprefix}{arXiv:}
\providecommand{\URLprefix}{URL: }
\providecommand{\Pubmedprefix}{pmid:}
\providecommand{\doi}[1]{\href{http://dx.doi.org/#1}{\path{#1}}}
\providecommand{\Pubmed}[1]{\href{pmid:#1}{\path{#1}}}
\providecommand{\bibinfo}[2]{#2}
\ifx\xfnm\relax \def\xfnm[#1]{\unskip,\space#1}\fi
\bibitem[{Ponti~Jr(2011)}]{ponti2011combining}
\bibinfo{author}{M.~P. Ponti~Jr},
\newblock \bibinfo{title}{Combining classifiers: from the creation of ensembles
  to the decision fusion},
\newblock in: \bibinfo{booktitle}{2011 24th SIBGRAPI Conference on Graphics,
  Patterns, and Images Tutorials}, \bibinfo{organization}{IEEE},
  \bibinfo{year}{2011}, pp. \bibinfo{pages}{1--10}.
\bibitem[{Kuncheva(2014)}]{kuncheva14}
\bibinfo{author}{L.~Kuncheva}, \bibinfo{title}{Combining Pattern Classifiers:
  Methods and Algorithms}, \bibinfo{edition}{2} ed., \bibinfo{publisher}{John
  Wiley and Sons}, \bibinfo{year}{2014}.
\bibitem[{Polikar(2012)}]{polikar12chapter}
\bibinfo{author}{R.~Polikar},
\newblock \bibinfo{title}{{Ensemble Learning}},
\newblock in: \bibinfo{editor}{C.~Zhang}, \bibinfo{editor}{Y.~Ma} (Eds.),
  \bibinfo{booktitle}{{Ensemble Machine Learning: Methods and Applications}},
  \bibinfo{publisher}{Springer}, \bibinfo{year}{2012}, pp.
  \bibinfo{pages}{1--34}. \DOIprefix\doi{10.1007/978-1-4419-9326-7\_1}.
\bibitem[{Breiman(1996)}]{breiman96}
\bibinfo{author}{L.~Breiman},
\newblock \bibinfo{title}{Bagging predictors},
\newblock \bibinfo{journal}{Machine Learning} \bibinfo{volume}{24}
  (\bibinfo{year}{1996}) \bibinfo{pages}{123--140}.
  \DOIprefix\doi{10.1023/A:1018054314350}.
\bibitem[{Ho(1998)}]{Ho98}
\bibinfo{author}{T.~K. Ho},
\newblock \bibinfo{title}{The random subspace method for constructing decision
  forests},
\newblock \bibinfo{journal}{IEEE Transactions on Pattern Analysis and Machine
  Intelligence} \bibinfo{volume}{20} (\bibinfo{year}{1998})
  \bibinfo{pages}{832--844}.
\bibitem[{Zhang and Ma(2012)}]{zhang12}
\bibinfo{editor}{C.~Zhang}, \bibinfo{editor}{Y.~Ma} (Eds.),
  \bibinfo{title}{Ensemble Machine Learning: Methods and Applications},
  \bibinfo{publisher}{Springer}, \bibinfo{year}{2012}.
  \DOIprefix\doi{10.1007/978-1-4419-9326-7}.
\bibitem[{G{\'e}ron(2019)}]{geron2019hands}
\bibinfo{author}{A.~G{\'e}ron}, \bibinfo{title}{Hands-On Machine Learning with
  Scikit-Learn, Keras, and TensorFlow: Concepts, Tools, and Techniques to Build
  Intelligent Systems}, \bibinfo{publisher}{O'Reilly Media},
  \bibinfo{year}{2019}.
\bibitem[{Breiman(2001)}]{breiman01}
\bibinfo{author}{L.~Breiman},
\newblock \bibinfo{title}{Random forests},
\newblock \bibinfo{journal}{Machine Learning} \bibinfo{volume}{45}
  (\bibinfo{year}{2001}) \bibinfo{pages}{5--32}.
  \DOIprefix\doi{10.1023/A:1010933404324}.
\bibitem[{Austin(1987)}]{austin87}
\bibinfo{author}{J.~Austin},
\newblock \bibinfo{title}{{ADAM: A Distributed Associative Memory for Scene
  Analysis}},
\newblock in: \bibinfo{booktitle}{{Proceedings of the IEEE First International
  Conference on Neural Networks}}, volume~\bibinfo{volume}{IV},
  \bibinfo{address}{San Diego}, \bibinfo{year}{1987}, p. \bibinfo{pages}{285}.
\bibitem[{Kohonen(1987)}]{kohonen87}
\bibinfo{author}{T.~Kohonen}, \bibinfo{title}{{Self-organization and
  associative memory}}, \bibinfo{edition}{2rd edition} ed.,
  \bibinfo{publisher}{Springer-Verlag New York, Inc.}, \bibinfo{address}{New
  York, NY, USA}, \bibinfo{year}{1987}.
\bibitem[{Hassoun and Watta(1997)}]{hassoun97}
\bibinfo{author}{M.~H. Hassoun}, \bibinfo{author}{P.~B. Watta},
\newblock \bibinfo{title}{{Associative Memory Networks}},
\newblock in: \bibinfo{editor}{E.~Fiesler}, \bibinfo{editor}{R.~Beale} (Eds.),
  \bibinfo{booktitle}{{Handbook of Neural Computation}},
  \bibinfo{publisher}{Oxford University Press}, \bibinfo{year}{1997}, pp.
  \bibinfo{pages}{C1.3:1--C1.3:14}.
\bibitem[{Hopfield(1982)}]{hopfield82}
\bibinfo{author}{J.~J. Hopfield},
\newblock \bibinfo{title}{{Neural Networks and Physical Systems with Emergent
  Collective Computational Abilities}},
\newblock \bibinfo{journal}{Proceedings of the National Academy of Sciences}
  \bibinfo{volume}{79} (\bibinfo{year}{1982}) \bibinfo{pages}{2554--2558}.
\bibitem[{Hopfield and Tank(1985)}]{hopfield85}
\bibinfo{author}{J.~Hopfield}, \bibinfo{author}{D.~Tank},
\newblock \bibinfo{title}{{Neural computation of decisions in optimization
  problems}},
\newblock \bibinfo{journal}{Biological Cybernetics} \bibinfo{volume}{52}
  (\bibinfo{year}{1985}) \bibinfo{pages}{141--152}.
\bibitem[{Smith et~al.(1998)Smith, Palaniswami, and Krishnamoorthy}]{smith98}
\bibinfo{author}{K.~Smith}, \bibinfo{author}{M.~Palaniswami},
  \bibinfo{author}{M.~Krishnamoorthy},
\newblock \bibinfo{title}{{Neural Techniques for Combinatorial Optimization
  with Applications}},
\newblock \bibinfo{journal}{IEEE Transactions on Neural Networks}
  \bibinfo{volume}{9} (\bibinfo{year}{1998}) \bibinfo{pages}{1301--1318}.
\bibitem[{Sun(2000)}]{sun00}
\bibinfo{author}{Y.~Sun},
\newblock \bibinfo{title}{{Hopfield neural network based algorithms for image
  restoration and reconstruction. II. Performance analysis}},
\newblock \bibinfo{journal}{IEEE Transactions on Signal Processing}
  \bibinfo{volume}{48} (\bibinfo{year}{2000}) \bibinfo{pages}{2119--2131}.
  \DOIprefix\doi{10.1109/78.847795}.
\bibitem[{Serpen(2008)}]{serpen08}
\bibinfo{author}{G.~Serpen},
\newblock \bibinfo{title}{{Hopfield Network as Static Optimizer: Learning the
  Weights and Eliminating the Guesswork.}},
\newblock \bibinfo{journal}{Neural Processing Letters} \bibinfo{volume}{27}
  (\bibinfo{year}{2008}) \bibinfo{pages}{1--15}.
  \DOIprefix\doi{10.1007/s11063-007-9055-8}.
\bibitem[{McEliece et~al.(1987)McEliece, Posner, Rodemich, and
  Venkatesh}]{mceliece87}
\bibinfo{author}{R.~J. McEliece}, \bibinfo{author}{E.~C. Posner},
  \bibinfo{author}{E.~R. Rodemich}, \bibinfo{author}{S.~Venkatesh},
\newblock \bibinfo{title}{{The capacity of the {Hopfield} associative memory}},
\newblock \bibinfo{journal}{IEEE Transactions on Information Theory}
  \bibinfo{volume}{1} (\bibinfo{year}{1987}) \bibinfo{pages}{33--45}.
\bibitem[{Kanter and Sompolinsky(1987)}]{kanter87}
\bibinfo{author}{I.~Kanter}, \bibinfo{author}{H.~Sompolinsky},
\newblock \bibinfo{title}{{Associative Recall of Memory without Errors}},
\newblock \bibinfo{journal}{Physical Review} \bibinfo{volume}{35}
  (\bibinfo{year}{1987}) \bibinfo{pages}{380--392}.
\bibitem[{M{\"u}ezzino{\v g}lu et~al.(2005)M{\"u}ezzino{\v g}lu, G{\"u}zelis,
  and Zurada}]{muezzinoglu05}
\bibinfo{author}{M.~M{\"u}ezzino{\v g}lu}, \bibinfo{author}{C.~G{\"u}zelis},
  \bibinfo{author}{J.~Zurada},
\newblock \bibinfo{title}{{An Energy Function-Based Design Method for Discrete
  {Hopfield} Associative Memory With Attractive Fixed Points}},
\newblock \bibinfo{journal}{IEEE Transactions on Neural Networks}
  \bibinfo{volume}{16} (\bibinfo{year}{2005}) \bibinfo{pages}{370--378}.
\bibitem[{Chiueh and Goodman(1991)}]{chiueh91}
\bibinfo{author}{T.~Chiueh}, \bibinfo{author}{R.~Goodman},
\newblock \bibinfo{title}{{Recurrent Correlation Associative Memories}},
\newblock \bibinfo{journal}{IEEE Trans. on Neural Networks} \bibinfo{volume}{2}
  (\bibinfo{year}{1991}) \bibinfo{pages}{275--284}.
\bibitem[{Garc\'{\i}a and Moreno(2004{\natexlab{a}})}]{garcia04a}
\bibinfo{author}{C.~Garc\'{\i}a}, \bibinfo{author}{J.~A. Moreno},
\newblock \bibinfo{title}{{The Hopfield Associative Memory Network: Improving
  Performance with the Kernel ``Trick''}},
\newblock in: \bibinfo{booktitle}{{Lecture Notes in Artificial Inteligence -
  Proceedings of IBERAMIA 2004}}, volume \bibinfo{volume}{3315} of
  \textit{\bibinfo{series}{{Advances in Artificial Intelligence -- IBERAMIA
  2004}}}, \bibinfo{publisher}{Springer-Verlag},
  \bibinfo{year}{2004}{\natexlab{a}}, pp. \bibinfo{pages}{871--880}.
\bibitem[{Garc\'{\i}a and Moreno(2004{\natexlab{b}})}]{garcia04b}
\bibinfo{author}{C.~Garc\'{\i}a}, \bibinfo{author}{J.~A. Moreno},
\newblock \bibinfo{title}{{The Kernel Hopfield Memory Network}},
\newblock in: \bibinfo{editor}{P.~M.~A. Sloot}, \bibinfo{editor}{B.~Chopard},
  \bibinfo{editor}{A.~G. Hoekstra} (Eds.), \bibinfo{booktitle}{{Cellular
  Automata}}, \bibinfo{publisher}{Springer Berlin Heidelberg},
  \bibinfo{address}{Berlin, Heidelberg}, \bibinfo{year}{2004}{\natexlab{b}},
  pp. \bibinfo{pages}{755--764}.
\bibitem[{Perfetti and Ricci(2008)}]{perfetti08}
\bibinfo{author}{R.~Perfetti}, \bibinfo{author}{E.~Ricci},
\newblock \bibinfo{title}{{Recurrent correlation associative memories: A
  feature space perspective}},
\newblock \bibinfo{journal}{IEEE Transactions on Neural Networks}
  \bibinfo{volume}{19} (\bibinfo{year}{2008}) \bibinfo{pages}{333--345}.
\bibitem[{Kultur et~al.(2009)Kultur, Turhan, and Bener}]{kultur2009ensemble}
\bibinfo{author}{Y.~Kultur}, \bibinfo{author}{B.~Turhan},
  \bibinfo{author}{A.~Bener},
\newblock \bibinfo{title}{Ensemble of neural networks with associative memory
  (enna) for estimating software development costs},
\newblock \bibinfo{journal}{Knowledge-Based Systems} \bibinfo{volume}{22}
  (\bibinfo{year}{2009}) \bibinfo{pages}{395--402}.
\bibitem[{Hancock and Pelillo(1998)}]{hancock98}
\bibinfo{author}{E.~R. Hancock}, \bibinfo{author}{M.~Pelillo},
\newblock \bibinfo{title}{{A Bayesian interpretation for the exponential
  correlation associative memory}},
\newblock \bibinfo{journal}{Pattern Recognition Letters} \bibinfo{volume}{19}
  (\bibinfo{year}{1998}) \bibinfo{pages}{149--159}.
\bibitem[{Kittler and Roli(2003)}]{kittler2003multiple}
\bibinfo{author}{J.~Kittler}, \bibinfo{author}{F.~Roli},
  \bibinfo{title}{Multiple Classifier Systems: First International Workshop,
  MCS 2000 Cagliari, Italy, June 21-23, 2000 Proceedings},
  \bibinfo{publisher}{Springer}, \bibinfo{year}{2003}.
\bibitem[{Hansen and Salamon(1990)}]{hansen1990neural}
\bibinfo{author}{L.~K. Hansen}, \bibinfo{author}{P.~Salamon},
\newblock \bibinfo{title}{Neural network ensembles},
\newblock \bibinfo{journal}{IEEE transactions on pattern analysis and machine
  intelligence} \bibinfo{volume}{12} (\bibinfo{year}{1990})
  \bibinfo{pages}{993--1001}.
\bibitem[{Van~Erp et~al.(2002)Van~Erp, Vuurpijl, and
  Schomaker}]{van2002overview}
\bibinfo{author}{M.~Van~Erp}, \bibinfo{author}{L.~Vuurpijl},
  \bibinfo{author}{L.~Schomaker},
\newblock \bibinfo{title}{An overview and comparison of voting methods for
  pattern recognition},
\newblock in: \bibinfo{booktitle}{Proceedings Eighth International Workshop on
  Frontiers in Handwriting Recognition}, \bibinfo{organization}{IEEE},
  \bibinfo{year}{2002}, pp. \bibinfo{pages}{195--200}.
\bibitem[{Ferreira and Figueiredo(2012)}]{ferreira12chapter}
\bibinfo{author}{A.~Ferreira}, \bibinfo{author}{M.~Figueiredo},
\newblock \bibinfo{title}{{Boosting Algorithms: A Review of Methods, Theory,
  and Applications}},
\newblock in: \bibinfo{editor}{C.~Zhang}, \bibinfo{editor}{Y.~Ma} (Eds.),
  \bibinfo{booktitle}{{Ensemble Machine Learning: Methods and Applications}},
  \bibinfo{publisher}{Springer}, \bibinfo{year}{2012}, pp.
  \bibinfo{pages}{35--85}. \DOIprefix\doi{10.1007/978-1-4419-9326-7\_2}.
\bibitem[{Pedregosa et~al.(2011)Pedregosa, Varoquaux, Gramfort, Michel,
  Thirion, Grisel, Blondel, Prettenhofer, Weiss, Dubourg, Vanderplas, Passos,
  Cournapeau, Brucher, Perrot, and Duchesnay}]{scikit-learn}
\bibinfo{author}{F.~Pedregosa}, \bibinfo{author}{G.~Varoquaux},
  \bibinfo{author}{A.~Gramfort}, \bibinfo{author}{V.~Michel},
  \bibinfo{author}{B.~Thirion}, \bibinfo{author}{O.~Grisel},
  \bibinfo{author}{M.~Blondel}, \bibinfo{author}{P.~Prettenhofer},
  \bibinfo{author}{R.~Weiss}, \bibinfo{author}{V.~Dubourg},
  \bibinfo{author}{J.~Vanderplas}, \bibinfo{author}{A.~Passos},
  \bibinfo{author}{D.~Cournapeau}, \bibinfo{author}{M.~Brucher},
  \bibinfo{author}{M.~Perrot}, \bibinfo{author}{E.~Duchesnay},
\newblock \bibinfo{title}{Scikit-learn: Machine learning in {P}ython},
\newblock \bibinfo{journal}{Journal of Machine Learning Research}
  \bibinfo{volume}{12} (\bibinfo{year}{2011}) \bibinfo{pages}{2825--2830}.
\bibitem[{Vanschoren et~al.(2013)Vanschoren, van Rijn, Bischl, and
  Torgo}]{OpenML}
\bibinfo{author}{J.~Vanschoren}, \bibinfo{author}{J.~N. van Rijn},
  \bibinfo{author}{B.~Bischl}, \bibinfo{author}{L.~Torgo},
\newblock \bibinfo{title}{Openml: Networked science in machine learning},
\newblock \bibinfo{journal}{SIGKDD Explorations} \bibinfo{volume}{15}
  (\bibinfo{year}{2013}) \bibinfo{pages}{49--60}.
  \DOIprefix\doi{10.1145/2641190.2641198}.
\bibitem[{Dem{\v s}ar(2006)}]{Demsar06}
\bibinfo{author}{J.~Dem{\v s}ar},
\newblock \bibinfo{title}{{Statistical comparisons of classifiers over multiple
  data sets}},
\newblock \bibinfo{journal}{Journal of Machine Learning Research}
  \bibinfo{volume}{7} (\bibinfo{year}{2006}) \bibinfo{pages}{1--30}.
\bibitem[{Burda(2013)}]{burda13}
\bibinfo{author}{M.~Burda}, \bibinfo{title}{{paircompviz: An {R} Package for
  Visualization of Multiple Pairwise Comparison Test Results}},
  \bibinfo{year}{2013}. \DOIprefix\doi{10.18129/B9.bioc.paircompviz}.
\bibitem[{Weise and Chiong(2015)}]{weise15}
\bibinfo{author}{T.~Weise}, \bibinfo{author}{R.~Chiong},
\newblock \bibinfo{title}{{An alternative way of presenting statistical test
  results when evaluating the performance of stochastic approaches}},
\newblock \bibinfo{journal}{Neurocomputing} \bibinfo{volume}{147}
  (\bibinfo{year}{2015}) \bibinfo{pages}{235--238}.
  \DOIprefix\doi{10.1016/j.neucom.2014.06.071}.
\bibitem[{Jankowski et~al.(1996)Jankowski, Lozowski, and Zurada}]{jankowski96}
\bibinfo{author}{S.~Jankowski}, \bibinfo{author}{A.~Lozowski},
  \bibinfo{author}{J.~Zurada},
\newblock \bibinfo{title}{{Complex-Valued Multi-State Neural Associative
  Memory}},
\newblock \bibinfo{journal}{IEEE Transactions on Neural Networks}
  \bibinfo{volume}{7} (\bibinfo{year}{1996}) \bibinfo{pages}{1491--1496}.
\bibitem[{M{\"u}ezzino{\v g}lu et~al.(2003)M{\"u}ezzino{\v g}lu, G{\"u}zeli{\c
  s}, and Zurada}]{muezzinoglu03}
\bibinfo{author}{M.~M{\"u}ezzino{\v g}lu}, \bibinfo{author}{C.~G{\"u}zeli{\c
  s}}, \bibinfo{author}{J.~Zurada},
\newblock \bibinfo{title}{{A New Design Method for the Complex-Valued
  Multistate {Hopfield} Associative Memory}},
\newblock \bibinfo{journal}{IEEE Transactions on Neural Networks}
  \bibinfo{volume}{14} (\bibinfo{year}{2003}) \bibinfo{pages}{891--899}.
\bibitem[{Minemoto et~al.(2016)Minemoto, Isokawa, Nishimura, and
  Matsui}]{minemoto16}
\bibinfo{author}{T.~Minemoto}, \bibinfo{author}{T.~Isokawa},
  \bibinfo{author}{H.~Nishimura}, \bibinfo{author}{N.~Matsui},
\newblock \bibinfo{title}{{Quaternionic multistate Hopfield neural network with
  extended projection rule}},
\newblock \bibinfo{journal}{Artificial Life and Robotics} \bibinfo{volume}{21}
  (\bibinfo{year}{2016}) \bibinfo{pages}{106--111}.
  \DOIprefix\doi{10.1007/s10015-015-0247-4}.
\bibitem[{Kobayashi(2017)}]{kobayashi17a}
\bibinfo{author}{M.~Kobayashi},
\newblock \bibinfo{title}{{Quaternionic Hopfield neural networks with
  twin-multistate activation function}},
\newblock \bibinfo{journal}{Neurocomputing} \bibinfo{volume}{267}
  (\bibinfo{year}{2017}) \bibinfo{pages}{304--310}.
  \DOIprefix\doi{10.1016/j.neucom.2017.06.013}.

\end{thebibliography}

\end{document}